\documentclass[11pt, a4paper, copyright, goog, gr]{google}

\usepackage[authoryear, sort&compress, round]{natbib}

\usepackage[most]{tcolorbox}

\newtcblisting{codebox}[1][]{%
  listing only,
  colback=gray!8,          
  colframe=gray!30,         
  arc=3pt,                  
  boxrule=0.6pt,            
  left=8pt, right=8pt,
  top=6pt, bottom=6pt,
  breakable,                
  listing engine=listings,
  listing options={
    basicstyle=\small\ttfamily, 
    breaklines=true,            
    columns=fullflexible,
    keepspaces=true
  },
  #1
}

\usepackage{amsmath, amssymb, amsthm}
\usepackage{booktabs}
\usepackage{xcolor}
\usepackage{tikz}
\usetikzlibrary{arrows.meta, positioning, shapes.geometric, fit, calc}

\newtheorem{theorem}{Theorem}

\newcommand{\cogentic}{Cogentic}

\keywords{multi-agent orchestration, automated proof discovery, LLM reasoning,
adversarial verification, inference efficiency}

\uselogo{}

\title{Cogentic: Multi-Agent Orchestration for Automated Proof Discovery}

\renewcommand{\today}{2026-10-01}

\author{Yang Cai}
\author{Vineet Gupta}
\author{Yanchen Jiang}
\author{Christopher Liaw}
\author{Aranyak Mehta}
\author{Grigoris Velegkas}
\author{Di~Wang}
\affil{Google Research}
\begin{abstract}
We present \cogentic{}, a multi-agent harness for automated proof discovery on
open research problems. While frontier language models can generate strong
mathematical ideas in a single shot, single-shot generation is often
insufficient for open problems that require exploring multiple competing
conjectures, overcoming subtle technical obstructions, and retaining
intermediate progress over a long horizon. \cogentic{} addresses these
challenges through an iterative prove--verify loop in which an orchestrator
allocates a population of independent provers across distinct proof directions,
subjects their output to adversarial verification by several specialized
components, and promotes confirmed intermediate results into a persistent verified ledger that later rounds build on. 
The harness is designed to be able to solve research-level math and theoretical computer science problems.
Using either Gemini 3.1 Pro or an early version of Gemini 4 Argon as the base model, \cogentic{} produced novel results on open problems
across online learning, auction theory, and mechanism design.
Each result was independently verified by domain
experts and is developed in full in companion papers. We list these results,
and new ones as they are verified, at
\url{https://sites.google.com/view/cogentic}.
\end{abstract}

\begin{document}

\maketitle

\renewcommand{\thefootnote}{}%
\footnotetext{Authors are listed in alphabetical order. The following authors have additional affiliations beyond Google
Research: Yang Cai (Yale University) and Vineet Gupta (Google DeepMind).}%
\renewcommand{\thefootnote}{\arabic{footnote}}

\section{Introduction}
\label{sec:intro}

We introduce \cogentic{}, a multi-agent harness for proof discovery on open
research problems. Inspired by our own experience conducting research in
theoretical computer science and mathematics, the harness divides the work like a research group would: an orchestrator decides what gets worked on, provers
draft proofs in parallel, and verifiers read those drafts looking for potential issues. 
We found that our system is quite efficient: The results reported here were based on \cogentic{} runs (each powered entirely by either Gemini 3.1 Pro or an early version of Gemini 4 Argon) which used $O(100)$ model calls for most
problems, and $O(1000)$ for the hardest. The system has headroom for further optimization of the number of calls but also allows for naturally scaling up to solve harder problems. A key feature of the system is that it only needs a problem statement, without any expert hints, and can work autonomously until it produces a result in the form of a paper.
To evaluate the harness, we focus on problem domains within our own areas of expertise, targeting natural-language, human-readable proofs where we can directly verify the mathematical reasoning and provide exposition surrounding the background of the paper and the novel techniques that the system generates.

Interactive theorem provers --- Lean \citep{moura2021lean4}, Isabelle/HOL
\citep{nipkow2002isabelle}, Coq \citep{bertot2013interactive} --- provide machine-checked guarantees, and a substantial line of work uses language models to lower the cost of obtaining them: retrieving premises and generating tactics \citep{yang2023leandojo}, drafting an informal proof and compiling it into a formal sketch \citep{jiang2022draft}, and training for olympiad-level formal reasoning with reinforcement learning \citep{hubert2026olympiad}. More
recently the same machinery has been used on open research problems \citep{tsoukalas2026advancingmathematicsresearchaidriven}.
In contrast, \cogentic{} operates in natural language, producing
mathematical prose that domain experts can verify.

A distinct and highly successful line of work uses language models to search for
mathematical objects rather than for arguments. Program search has discovered
new constructions in extremal combinatorics \citep{romera2024mathematical},
evolutionary coding agents have improved algorithms and bounds
\citep{alphaevolve}, and related systems perform test-time learning on open
problems \citep{wang2025thetaevolve, yuksekgonul2026learning} or explore many
problems at once \citep{georgiev2025mathematical}.
Neuro-symbolic search
attains medal-level performance in olympiad geometry by pairing a learned
proposer with a symbolic engine \citep{trinh2024solving}. The same recipe
extends beyond mathematics to writing expert-level empirical research software
under a quality metric \citep{aygün2026aihelpscientistswrite}. What each
of these approaches requires is a cheap, faithful, and machine-computable score.
Sometimes theorems can be proven by reducing the question to finding mathematical structures with certain properties: e.g., \cite{nagda2026reinforcedgenerationcombinatorialstructures} found proofs of inapproximability for combinatorial problems by finding extremal structures, and  \citep{cai_et_al} found bounds on gain from bilateral trade by finding extremal probability distributions, both using AlphaEvolve~\citep{alphaevolve} with such oracles.
However, not every problem is amenable to such an approach.

A second line of work is based on scaling the LLM inference budget, perhaps through multiple model calls, to solve more complex tasks.
The base case is simply to sample more --- chain-of-thought
prompting \citep{wei2022chain}, self-consistency across sampled solutions
\citep{wang2022self}, and repeated sampling with a selector over the
candidates \citep{brown2024large, snell2024scaling}.
Another approach is to scale the inference via multi-agent interaction.
Debate between instances has been reported to improve
factuality and reasoning \citep{du2023improving, liang2023encouraging}, and
models have been used to evaluate the work of other models
\citep{zhuge2024agent}. 
Our framework builds on an iterative interaction between provers and verifiers.
The verifiers are adversarial and begin with the assumption that the proofs are incorrect or incomplete.
We explain our framework in more detail in Section~\ref{sec:system}.

There are several recent works that apply agentic harnesses to
research-level mathematics and theoretical computer science. We do not attempt to give a comprehensive survey or compare these efforts, but they include
\citep{feng2026towards,lin2026colosseum,schmitt2026proofcouncilllmagentsolving,zheng2026aicomathematicianacceleratingmathematicians,Gottweis_2026}.

Finally, we place our contributions against recent milestones obtained with
large agentic systems, including the resolution of a Millennium Prize problem \citep{OAI_millennium} and other long-standing mathematical conjectures \citep{OAI_conjectures, Ant}. We focus on problems in our areas of expertise, at the hardness level of open questions in flagship theoretical computer science conferences like STOC and FOCS, using a relatively low inference budget, $O(100)$ to $O(1000)$ Gemini calls per
problem.

\section{The Harness}
\label{sec:system}

\cogentic{} consists of a group of agents working on a single problem.
They share a workspace on disk and are coordinated by an orchestrator that decides
which of them runs, when, and on what.

\paragraph{Components.}
We outline the main components of our system below.
\begin{itemize}
\item The \emph{orchestrator} serves as the central controller: it tracks global
state, partitions prover slots across research directions, spawns summarizers to condense history into targeted briefings for individual provers, evaluates verifier consensus, and manages ledgers that store important findings obtained so far and records that track prior proof attempts and their verdicts. The orchestrator does not perform mathematical derivations itself.
\item \emph{Literature reviewers} search for related work and retrieve information such as definitions and theorems that a prover is likely to need.
They can also be sent out again mid-run to do a more targeted search and find results that can help overcome ongoing technical barriers.
\item \emph{Provers} generate candidate proofs independently and in parallel based on
assigned briefings that selectively summarize findings obtained so far. \item \emph{Verifiers} critique candidate solutions from complementary angles and
scopes. 
\item The \emph{record} keeps track of prover attempts and their corresponding critiques from verifiers, and the \emph{ledger} keeps track of verified intermediate lemmas that came out of proof attempts. These are the central ways that our agents communicate and document progress, as explained in more detail below.

\item An \emph{advisor} reads outputs across rounds to help the orchestrator decide how to control and allocate prover attempts and what to adjust the instructions given to each individual prover for the next round.
\item Finally, a \emph{consolidation stage} formats and audits the final manuscript.
\end{itemize}

\begin{figure}[t]
\centering
\includegraphics[width=\textwidth]{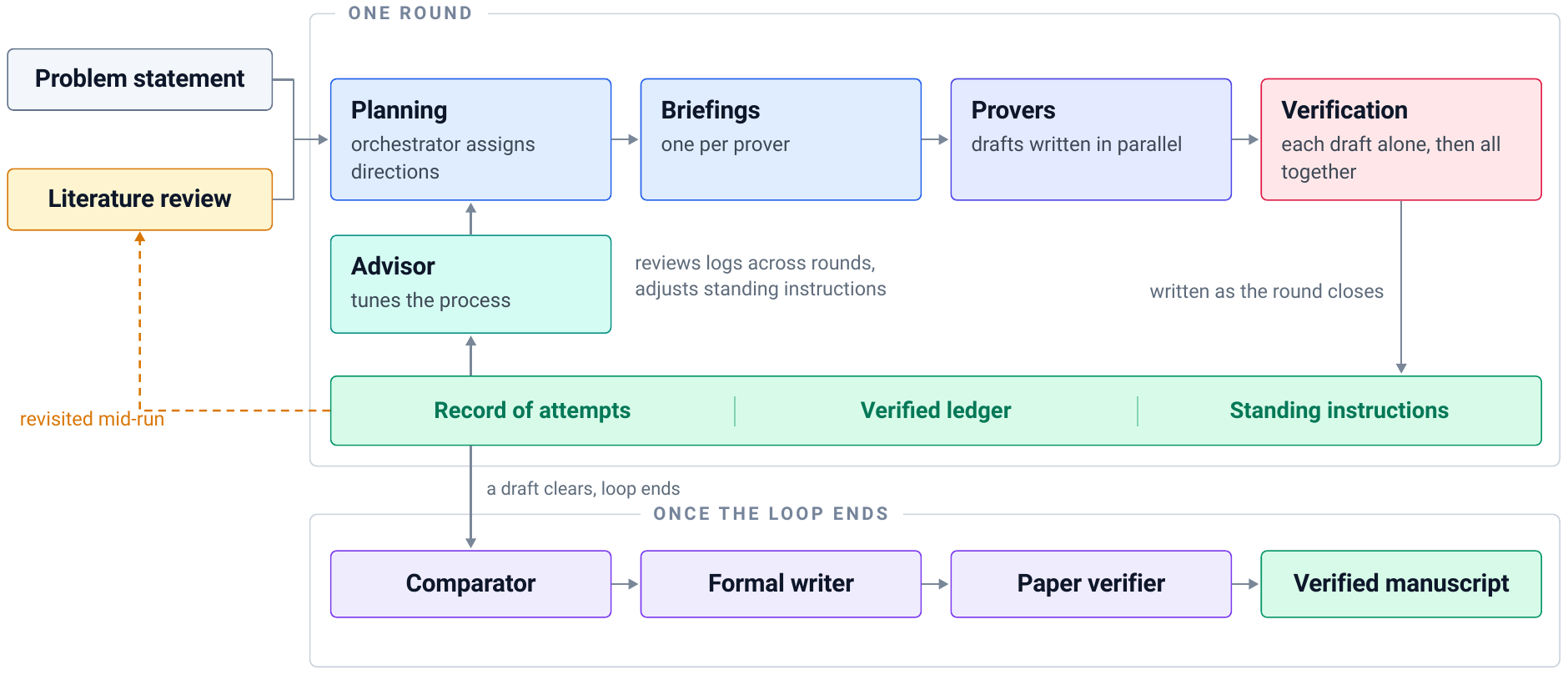}
\caption{One round of \cogentic{}. The orchestrator decides how many provers to
run and which direction each attends to, an advisor writes each prover an
individual briefing, the provers draft candidate proofs in parallel, and every
draft is verified both on its own and alongside the others from the round. What
a round establishes is written down for subsequent rounds: the attempts and why
they failed, the verified ledger of intermediate results, and the standing
instructions maintained by the process advisor. Literature reviewers supply
background at the start and can be dispatched again mid-run when attempts stall
at the same step. Rounds continue until a draft clears verification after which the accepted proof is consolidated,
expanded into a formal manuscript, and audited.}
\label{fig:overview}
\end{figure}

\paragraph{Workflow.} The work proceeds in rounds (Figure~\ref{fig:overview}). A round produces a batch
of candidate proofs, puts them through verification, and writes into the record and the verified ledger that the next round starts from.
Rounds continue
until a draft clears verification or
until the budget runs out.

\subsection{Directions and Assignments}

A round opens with the orchestrator planning the work: how many provers to run,
and what direction each should attend to. A direction here is a specific claim (e.g., a bound, a constant, a construction), finding counterexamples, or as the run goes on, repairing and finalizing promising proofs.
A prover is assigned to a direction to work on, but not
how to work on it. It reads what has been attempted previously for that direction, what the verifiers
said about those attempts, as well as common information like the literature survey and the verified ledger, and chooses its own next step. 

As the run goes on, the material relevant to a prover grows in volume. Rather than passing all of it as input, each prover sees only a briefing, written for it by a summarizer, which is spawned by the orchestrator. The summarizer reads all prior attempts and their verdicts, selects the most informative ones as context, and gives suggestions for possible next steps,
based on what has worked and what has not. Each summarizer produces its briefing independently, so provers in the same round receive different readings of the same history.
Longer documents are
given as paths rather than quoted in full, so a prover (which itself is an agent, such as an Antigravity agent~\citep{google2025antigravity}) can open what it wants to
read selectively.

\subsection{Verification}
\label{sec:verification}
After each prover completes a proof, it is read by a verifier whose task is to check the correctness of the proof. A separate verifier reads all of the round's drafts side by side, which allows it to spot shared blind spots and compare the strengths of individual proofs. A draft is accepted only when it passes both verifiers. Both are adversarial: they start from the assumption that every step is wrong until justified, and that every citation is incorrect until checked.

\subsection{Cross-Round Communication}
\label{sec:ledger}
 Each round leaves two artifacts: a record of what was tried, and a ledger of results that have been verified. The \emph{record} pools attempts by what they were trying to establish, storing with each one the method used and the objection it failed on. The orchestrator and advisor read the record and decide what the next round should do.

    The \textit{ledger} accumulates confirmed fragments, such as intermediate lemmas, across rounds. Verifiers often confirm individual lemmas inside proofs they otherwise reject. At the end of a round, an auditor extracts those fragments, rewrites each as a self-contained lemma, and sends it to be verified again in isolation. What survives becomes available to every later prover as something that may be reused without being proved again. The ledger also records dead ends, for example, a bound excluded by a verified counterexample is written down as excluded, so that later rounds do not revisit them.

\subsection{Process Level Adaptation}
At the end of each round, a process advisor reviews verification logs, not only from the current round but from the run as a whole, to adjust how subsequent rounds are conducted. It surfaces patterns that become visible over time: repeated mistakes, common gaps in arguments, recurring verification blind spots where one verifier missed but was caught by some other verifier, etc. Based on these observations, it recommends adjustments to the instructions given to provers and verifiers — for example, a warning about a mistake several provers keep making, or a stricter justification standard for the kind of step where earlier drafts cut corners. It also advises the orchestrator on how to allocate prover attempts and how to adjust the instructions given to each prover and verifier for the next round. This is a separate agent because the orchestrator, which must juggle concurrent duties across many agents, rarely has the context window to review and synthesize log trends itself. Note that, neither the advisor nor the orchestrator is permitted a mathematical opinion: they cannot speculate on what the answer is likely to be, recommend a technique, or declare a direction promising or dead.

\subsection{Termination and Output}
\label{sec:termination}

When a candidate clears all verification, the orchestrator may continue to explore some remaining promising that may yield a better result, terminating once those avenues are
exhausted. Upon termination, a comparator selects the strongest verified proof, a
formal writer expands it into a complete manuscript, and a final verification
audit checks the compiled document against the accepted proof to ensure no errors
were introduced during exposition. The output is a self-contained document that states the theorems and the proofs, with the lemmas it depends on, notation, and background of the problem written out in full, so that it can be read and checked by domain experts who knows nothing about the run that produced it.

\section{Results}
\label{sec:results}

We applied \cogentic{} using either Gemini 3.1 Pro or an early version of Gemini 4 Argon as the base model to open research problems across
online learning, auction theory, and mechanism design. The problems are in the areas of the research expertise of the authors. Each run operated from the problem statement without human
mathematical intervention; domain experts subsequently verified every proof.
Table~\ref{tab:results} summarizes the five results, which are developed in
full in companion papers. We maintain an up-to-date list of results obtained
with \cogentic{}, with links to their companion papers, at
\url{https://sites.google.com/view/cogentic}.

\begin{table}[t]
\centering
\small
\caption{Open problems resolved or improved by \cogentic{}. Each result is
developed in full in a companion paper; see
\url{https://sites.google.com/view/cogentic} for updates.}
\label{tab:results}
\begin{tabular}{@{}p{4.3cm}p{2.7cm}p{3.6cm}p{3.8cm}@{}}
\toprule
\textbf{Problem} & \textbf{Area} & \textbf{Prior State of the Art} & \textbf{\cogentic{} Result} \\
\midrule
Online inverse linear optimization / low-regret cutting planes & Online learning \& optimization & $O(d \ln T)$ efficiently; $O(d)$ only by an improper rule costing $T^{\Theta(d)}$ per round  & \textbf{First efficient and first proper $O(d)$ bound, uniform in $T$, at $O(d^2)$ per round} \citep{inverseopt2026} \\
\addlinespace
Two-sided Bulow--Klemperer competition complexity & Auction theory \& market design & $O(1)$ agents suffice, but recruiting on \emph{both} sides and with a constant of at least $20{,}000$ per side & \textbf{$+2$ agents on the smaller side alone suffice ($\operatorname{STR}(m,n+2) \ge \operatorname{OPT}(m,n)$), and $+1$ does not suffice for any DSIC, IR, weakly budget-balanced mechanism} \citep{bk2026} \\
\addlinespace
Anytime regret with $n$ experts & Online learning & Anytime $\sqrt{t \ln n}$ vs.\ fixed-horizon $\sqrt{t \ln n / 2}$; the factor $\sqrt{2}$ open & \textbf{Anytime regret $\bigl(1 + O(\sqrt{\ln\ln n / \ln n})\bigr)\sqrt{t \ln n / 2}$: no leading-order price for anytime validity} \citep{anytime2026} \\
\addlinespace
Simple vs.\ optimal revenue maximization, single additive buyer & Mechanism design & $5.2 \cdot \max(\operatorname{SRev}, \operatorname{BRev}) \ge \operatorname{OPT}$ \citep{ma2021reaping} & \textbf{$3.52 \cdot \max(\operatorname{SRev}, \operatorname{BRev}) \ge \operatorname{OPT}$} \citep{multiitem2026} \\
\addlinespace
Price of anarchy for autobidding auctions & Auction theory \& autobidding &  $1.8$ PoA for $2$ bidders; general $n$-bidder tight mechanism open & \textbf{Optimal $1.5$ PoA for $2$ bidders (anonymous, monotone mechanisms); $2 - \frac{1}{4n+1}$ for $n$ bidders} \citep{autobidding2026} \\
\bottomrule
\end{tabular}
\end{table}

\subsection{Efficient Online Inverse Linear Optimization}
\label{sec:inverseopt}

In online inverse linear optimization, a learner watches an expert make choices
and tries to make the same ones without ever being told what the expert is
optimizing. A vector $w^{*}$ in the unit ball $\mathbb{B} \subseteq
\mathbb{R}^{d}$ is fixed and hidden. At each round an adversary reveals a
compact action set $X_t \subseteq \mathbb{B}$, the learner recommends some
$\hat{x}_t \in X_t$, and then observes the expert's choice $x_t \in
\arg\max_{x \in X_t} \langle w^{*}, x \rangle$. It does not see $w^{*}$ or the
value of any action. The
learner is charged the cumulative shortfall $R_T = \sum_{t \le T} \langle
w^{*}, x_t - \hat{x}_t \rangle$, a sum of non-negative terms, and the question
is whether that sum can be bounded by a function of $d$ alone, uniformly in $T$.

A bound uniform in $T$ is stronger than
$O(\sqrt{T})$ or $O(d \ln T)$, both of which let the total grow without bound.
And a learner is \emph{proper} if it commits to a nonzero estimate $\hat{w}_t$ of the
objective before seeing the menu and then recommends a maximizer of $\langle
\hat{w}_t, \cdot \rangle$; a proper learner answers not only what to do but
what it takes the expert to be optimizing, and its recommendation costs one
linear optimization over $X_t$. The problem reduces to a cutting-plane game
with a strong separation oracle, in which the learner queries $p_t \in
\mathbb{R}^d$ and an adversary who has seen $p_t$ returns a unit vector $v_t$
with $\langle w^{*} - p_t, v_t \rangle \ge 0$; bounding the regret of that game suffices to
bounds $R_T$.

Prior work left a gap between the two desiderata. For arbitrary action sets the
best efficient bounds were $O(d \ln T)$, polynomial but growing with the
horizon: \citet{gollapudi2021contextual} obtained this from a regularized center of
gravity, and \citet{sakaue2026online} obtained it efficiently from an online
Newton step, at $O(d^2)$ per round after \citet{sakaue2026simple} removed the
Mahalanobis projection.
Bounds uniform in $T$ were either exponential in $d$ ---  the John
ellipsoid rule of \citet{gollapudi2021contextual}, at $\exp(O(d \log d))$ --- or
required extra structure on the action sets
\citep{sakaue2025revisiting,oki2026finite}. Whether a finite $\operatorname{poly}(d)$ bound was achievable at
all was answered affirmatively by \citet{dewasurendra2026multiscale}, but by a rule that is neither
proper nor efficient: it pools covers of the optimality-gap class across dyadic scales into a
weighted vote whose finite implementation can involve $T^{\Theta(d)}$
tests and requires nonconvex optimization over the action set, without
committing to a linear objective before seeing the menu.

\begin{theorem}[\citealp{inverseopt2026}]
\label{thm:inverseopt}
There is a deterministic, proper, anytime algorithm for online inverse linear
optimization with cumulative shortfall $R_T = O(d)$, uniform in the horizon
$T$, using $O(d^{2})$ arithmetic and one linear optimization per round.
\end{theorem}

This is the first bound of that order that is efficient, and the first that is
proper. It is within a factor of $O(\sqrt{d})$ of the optimal bound: every algorithm suffers
$\Omega(\sqrt{d})$ \citep{sakaue2026online}.

The proof builds on the variable-metric framework of \citet{sakaue2026online}, in which the learner runs online gradient descent under a metric $H_t$
that stretches along directions already queried. Two changes remove the
logarithm. First, with $g_t=-v_t$, the rank-one metric update $g_t g_t^{\top}$ is
\emph{self-normalized}, divided by the dual-metric length $s_t = \|g_t\|_{H_t^{-1}}$
of the observed direction. Second, the $\log\det H_t$ potential --- which
bounds the total squared step length $\sum_t s_t^2$ but grows like $d \log T$,
and is the source of the $\ln T$ in every earlier volumetric analysis --- is
replaced by the trace power $\operatorname{tr}(H_t^{-1/2})$. That potential starts at $d$, stays non-negative, and falls by at least
$\tfrac{\tau}{4}s_t^2$ each round, where $\tau = \Theta(1/d)$ controls
the metric update, so $\sum_t s_t^2 \le 4d/\tau = O(d^2)$, independently
of the horizon. With the iterate step size $\alpha = \Theta(1/d)$,
the distance-contraction argument and the reduction then give
$R_T = O(1/\alpha + \alpha\sum_t s_t^2) = O(d)$.

The argument uses the expert's optimality only to make each round legal, so the
same bound holds when the expert merely does at least as well as the learner by
its own criterion; this is the first $O(d)$ bound competing against an expert
that does not optimize. The companion paper also gives corruption-robust and rank-adaptive variants,
and an application to convex minimization: an $L$-Lipschitz convex function
with a minimizer within distance $R$ of the initial point can be minimized
from subgradient directions alone with total suboptimality $O(dLR)$ over
an infinite run.

Subsequent to the companion paper, \citet{sakaue2026tight} obtained a tight $O(\sqrt{d})$ bound. However, this algorithm is inefficient and it remains an open question to obtain an efficient algorith with $O(\sqrt{d})$ regret.

The base model used for this problem was Gemini 3.1 Pro.

\subsection{Competition Complexity for Two-Sided Markets}
\label{sec:bk}

The Bulow--Klemperer theorem \citep{bulow1996auctions} establishes that in
one-sided auctions, adding a single bidder to a simple mechanism yields at
least the revenue of the optimal mechanism for the original market. A natural
question is whether a Bulow--Klemperer-type result also holds for two-sided
markets. This line of work was initiated by \citet{babaioff2020bulow}, who show
that, when the buyers' valuations first-order stochastically dominate the
sellers' costs, adding roughly $n(m + 4\sqrt{m})$ buyers suffices for a
prior-independent mechanism to have gains from trade (GFT) at least that of the
first-best GFT, where $m$ is the number of buyers, $n$ the number of sellers,
and $m \ge n$. Following up on this, \citet{cai2024power} showed that if one
augments \emph{both} sides of the market then $O(1)$ agents suffice. Their
analysis, however, requires a large constant, namely at least $20{,}000$ per
side. At least two questions remained open, including (i) whether the constant can
be improved and (ii) whether it is sufficient to recruit from only one side of
the market.

Using \cogentic{} we proved the following theorem, which answers both questions
simultaneously.

\begin{theorem}[\citealp{bk2026}]
\label{thm:bk}
For any $m \ge n \ge 1$, if the buyer distribution $F_B$ first-order
stochastically dominates the seller distribution $F_S$, then adding exactly two
sellers --- the smaller side of the market --- suffices for Seller Trade
Reduction to achieve expected gains from trade at least the first-best gains
from trade of the original market:
\[
  \operatorname{STR}(m, n+2) \;\ge\; \operatorname{OPT}(m, n).
\]
Moreover this is tight: even for $m = n = 1$, recruiting one additional seller does not suffice for any
prior-independent mechanism.
\end{theorem}

The constant drops from at least $20{,}000$ per side to $2$ on one side, and
the recruitment is one-sided, so the mechanism designer need only find two more
participants of the type already in shorter supply.

The base model used for this problem was an early version of Gemini 4 Argon.

\subsection{Anytime Regret with $n$ Experts}
\label{sec:anytime-regret}

In prediction with expert advice, a learner plays a distribution over $n$
experts, an adversary reveals a loss vector in $[0,1]^n$, and the learner is
charged its expected loss relative to the best single expert in hindsight. When
the horizon $T$ is known, multiplicative weights tuned to $T$ guarantees regret
$\sqrt{T \ln n / 2}$, and the leading constant $1/\sqrt{2}$ cannot be improved
for large $n$ \citep{cesa1997use}. An \emph{anytime} algorithm is given
no horizon and must satisfy its bound at every $t$ simultaneously. The best
known anytime guarantee for many experts was $\sqrt{t \ln n}$, a factor
$\sqrt{2}$ worse, and whether that factor was necessary had remained open.
For $n = 2$, \citet{luo2014towards} proved that the anytime regret is strictly larger than the fixed-time regret obtained by \citet{cover1966behavior}.
Later, \citet{harvey2023optimal} gave the optimal anytime algorithm for $n = 2$ but also conjectured that as $n \to \infty$, the leading constant of the anytime regret and fixed-time regret would coincide.
Some evidence towards this conjecture was given by \citet{harvey2024continuous}, who proved that in continuous time, the anytime and fixed-time constants agree as $n \to \infty$ when the experts follow independent Brownian motions.

With the Cogentic framework, we prove that the leading constant of the fixed-time regret and anytime regret coincide as $n \to \infty$.
\begin{theorem}[\citealp{anytime2026}]
\label{thm:anytime}
There is an algorithm for prediction with expert advice which uses no knowledge
of the horizon and whose regret over $n$ experts satisfies
\[
  R_t \;\le\; \left(1 + O\!\left(\sqrt{\frac{\ln \ln n}{\ln n}}\right)\right)
  \sqrt{\frac{t \ln n}{2}}
  \qquad \text{simultaneously for all } t \ge 1
\]
for every sequence of loss vectors in $[0,1]^n$.
\end{theorem}

The leading constant is optimal, since an anytime algorithm is in particular a
fixed-horizon algorithm at every $T$.

The proof \cogentic{} found is elementary. Run one multiplicative-weights
instance for each horizon on a geometric grid $H^m = (1+\varepsilon)^m$ and
aggregate them with a master multiplicative-weights algorithm. At any time $t$
some instance is tuned to a horizon in $[t, (1+\varepsilon)t]$ and so has
regret $\sqrt{1+\varepsilon}\sqrt{t \ln n / 2}$; it would suffice for the
master to track that instance cheaply. The obstruction is that the pool grows
with $t$ while the master's regret grows with the number of instances it
tracks, which would leave a time-dependent overhead in the leading constant.
The construction keeps the pool bounded by waking instance $m$ only at round
$\lfloor \delta H^m \rfloor$ and retiring it after round $\lfloor H^m \rfloor$.
Then $O(\varepsilon^{-1} \log \delta^{-1})$ instances are awake at once,
independent of $t$, and at most that many are born during any one instance's
lifetime.
This second count matters because when new instances arrive, we must allocate some of the existing probability mass to new instances.
Retirement leaves
the prefix before $\lfloor \delta H^m \rfloor$ uncovered, and recursing on it
contributes a geometric series $1 + \sqrt{\delta} + \delta + \cdots$. Taking
$\varepsilon = \sqrt{\ln \ln n / \ln n}$ and $\delta = \varepsilon^3$ makes the
grid coarseness, the aggregation overhead, and the recursion together cost a
factor $1 + O(\varepsilon)$.

The base model used for this problem was an early version of Gemini 4 Argon.

\subsection{Simple versus Optimal Revenue for an Additive Buyer}
\label{sec:multiitem}

Consider a single buyer with additive valuations over $n$ items whose values
are drawn independently. Let $\operatorname{SRev}$ be the revenue from selling
each item separately at a fixed price, and $\operatorname{BRev}$ the revenue
from selling the grand bundle at a single price. Let $\operatorname{OPT}$ be the
revenue of the optimal mechanism, which may be randomized and arbitrarily
complex. \citet{babaioff2020simple} showed that the better of these two simple
mechanisms is within a constant factor of optimal:
$\operatorname{OPT} \le 6 \cdot \max(\operatorname{SRev}, \operatorname{BRev})$.
The duality framework of \citet{cai2016duality} later extended this result to a
more general setting. For the original single-additive-buyer setting,
\citet{ma2021reaping} improved the factor to $5.2$. The optimal constant is
still unknown, and the best lower bound on the approximation ratio is $2$~\citep{rubinstein2016}.

\begin{theorem}[\citealp{multiitem2026}]
\label{thm:multiitem}
For a single additive buyer whose values for $n$ items are independent,
\[
  3.52 \cdot \max(\operatorname{SRev}, \operatorname{BRev}) \;\ge\; \operatorname{OPT}.
\]
\end{theorem}

The proof builds on the duality framework of \citet{cai2016duality}. Let $v_j$ be
the buyer's value for item $j$, and let $V=\sum_j v_j$,
$M_{\max}=\max_j v_j$, and $M=\max\{\operatorname{SRev},\operatorname{BRev}\}$. The framework gives
\[
    \operatorname{OPT} \le \operatorname{SRev}+\mathbb{E}[V-M_{\max}],
\]
so it remains to bound the expected \emph{non-favorite welfare}
$\mathbb{E}[V-M_{\max}]$, i.e., the total value of all items except the
buyer's favorite.

Prior analyses truncate item values at $\operatorname{SRev}$ and split
$\mathbb{E}[V-M_{\max}]$ into a $\operatorname{Core}$ and a $\operatorname{Tail}$, each
bounded separately~\citep{cai2016duality}. We instead truncate at the joint scale $M$,
which also accounts for $\operatorname{BRev}$, and show directly that
\[
    \mathbb{E}[V-M_{\max}] \le \mathbb{E}[V_M],
    \qquad V_M=\sum_{j=1}^n \min\{v_j,M\}.
\]
This removes the $\operatorname{Tail}$ analysis entirely and leaves a single
$\operatorname{Core}$-like quantity. We then derive extremal distributions that maximize and minimize the second-order moment bound $\mathbb{E}[V_M^2]$, while keeping $\mathbb{E}[V_M]$ fixed. The upper and lower bounds then suggest a relationship between $\mathbb{E}[V_M]$ and $M$ that allows us to bound $\mathbb{E}[V_M]$ by a constant
multiple of $M$.

The base model used for this problem was Gemini 3.1 Pro.

\subsection{Price of Anarchy for Autobidding Auctions}
\label{sec:autobidding}

Automated bidding (autobidding) is now a widely adopted interface for advertisers to bid into high frequency ad auctions.
In this interface, advertisers specify high level constraints, such as return-on-spend and budget constraints.
To measure efficiency, the literature uses a standard metric known as the price of anarchy (PoA) which is the ratio of the welfare in the optimal allocation and the worst-case equilibrium welfare.
\citet{Aggarwal2019AutobiddingWC} initiated this line of work, showing that the second-price auction (SPA) achieves a tight PoA of $2$. Subsequent work by \citet{liaw2023efficiency} proved that no deterministic mechanism---including the first-price auction (FPA)---can beat the PoA barrier of $2$ in the prior-free setting, even for two bidders. When randomization is combined with non-truthful payments, however, the barrier of $2$ can be broken for two bidders: \citet{mehta2022autobidding} achieved a two-bidder PoA of approximately $1.89$ via a randomized truthful auction, and \citet{liaw2023efficiency} improved the two-bidder upper bound to $1.8$ using a randomized first-price auction, while proving an $n$-bidder \emph{lower bound} (for even $n$) showing that every anonymous mechanism has $\mathrm{PoA} \ge \frac{2n+4}{n+4} = 2 - \frac{4}{n+4}$. At least two questions remain open: (i) what is the exact minimax optimal PoA for $n = 2$ bidders, and (ii) for general $n \ge 3$ bidders, whether any mechanism can break the deterministic PoA barrier of $2$ and match the $2 - \Theta(1/n)$ lower bound.

Using Cogentic we proved the following theorem, which answers both questions simultaneously using the family of $r$-Proportional First-Price Auctions ($\mathsf{pFPA}_r$), in which each bidder $i \in [n]$ wins a query with probability $x_{i,j} = b_{i,j}^r / \sum_{k=1}^n b_{k,j}^r$ and pays their bid $b_{i,j}$ upon winning (a bidder facing no competing bid wins for free).

\begin{theorem}[\citealp{autobidding2026}]
In prior-free autobidding markets with return-on-spend constraints:
\begin{enumerate}
    \item For $n = 2$ bidders, the standard Proportional First-Price Auction ($\mathsf{pFPA}_1$, with $r=1$) achieves a Price of Anarchy of at most $1.5$. Moreover, this is tight: any anonymous two-bidder mechanism (with mild assumptions) has $\mathrm{PoA} \ge 1.5$.
    \item For general $n \ge 2$ bidders, the $2n$-Proportional First-Price Auction ($\mathsf{pFPA}_{2n}$, with $r=2n$) achieves a Price of Anarchy of at most
    \[
        \mathrm{PoA}(\mathsf{pFPA}_{2n}) \le 2 - \frac{1}{4n+1} = 2 - \Omega(1/n).
    \]
    This matches the $2 - \frac{4}{n+4}$ lower bound up to constant factors in the $1/n$ term.
\end{enumerate}
\end{theorem}
In the second part of the theorem, the upper bound holds only assuming that bids are undominated (i.e.~no bidder can raise their bid on any query to win more while satisfying their RoS constraint).

For the first part of this theorem, the authors had already suspected that the proportional first-price auction would have an improved Price of Anarchy over rFPA, 
and their conjecture was that the bound is $1.5$, 
but did not have a proof.
Cogentic was able to prove both the upper bound and the lower bound for any mechanism.
The authors had not previously studied the second part of the theorem and did not provide any hints to the system.
It independently came up with the mechanism and analysis.

The base model used for this problem was Gemini 3.1 Pro.

\section{Discussion}
\label{sec:discussion}

As noted, \cogentic{} produces natural language proofs which are verified by experts.
That was possible because of how the problems were chosen: they come from areas
the authors work in.
Some companion papers include coauthors who had already been working on the corresponding problems.
We checked the argument, wrote the exposition around it, and in some cases carried it further than the harness had. We note that the initial papers were coherent and nicely readable on their own, but we added further exposition such as better placement in the literature, the framing, and distilling and explaining the techniques.

A system like this can produce candidate results faster than they can be read,
and the gap widens as the compute budget grows. 
One possibility is to formalize in a proof assistant such as Lean~\citep{moura2021lean4}, so
that correctness is settled mechanically. However, human understanding of the solution might lag behind.
In the past, understanding the solution of a problem has also led to new directions and problems being explored.
Balancing out the throughput of this generation and human understanding remains an important question.

\bibliographystyle{plainnat}
\bibliography{references}

\appendix

\section{Prompts Used for the Problems}

\subsection{Efficient Online Inverse Linear Optimization}
\begin{codebox}
Let $B = \{w \in \mathbb{R}^d \mid \Vert{}w\Vert{}_2 \le 1\}$ be the unit ball in $\mathbb{R}^d$. We are searching for a hidden point $w^* \in B$. Every round $t$, we can choose a point $p_t \in B$ and submit this point to a separation oracle. The separation oracle then returns a half-space separating $p_t$ from $w^*$; in particular, the oracle returns a normalized direction $v_t$ (where $\Vert{}v_t\Vert{}_2 = 1$) such that $\langle w^*, v_t \rangle \ge \langle p_t, v_t \rangle$.
    
Traditionally, cutting-plane algorithms have been developed to minimize the number of calls to the separation oracle until the oracle returns a hyperplane that passes within some distance $\delta$ of $w^*$.
    
In our setting, instead of trying to minimize the number of separation oracle queries before finding a "close" hyperplane, we would like to minimize the total (over all $T$ rounds) distance between the returned hyperplanes and the hidden point $w^*$. That is, we would like to minimize the expression:$Reg' = \sum_{t=1}^T (\langle w^*, v_t \rangle - \langle p_t, v_t \rangle) = \sum_{t=1}^T \langle w^* - p_t, v_t \rangle$. 
    
Can we design a cutting plane algorithm whose regret is polynomial in $d$, and independent of $T$?
    
Note that this problem is from the paper NeurIPS 2021 paper "Contextual Recommendations and Low-Regret Cutting-Plane".
\end{codebox}

\subsection{Recruiting Two Traders Suffices for Two-Sided Markets}
\begin{codebox}
Read this paper: https://arxiv.org/abs/2307.03844.

Suppose there are $m$ buyers and $n$ sellers and $m \geq n$.
We assume that the buyer distribution $F_B$ first-order stochastically dominates the seller distribution $F_S$.

Let STR(m, n) be the expected GFT of Seller Trade Reduction as discussed in the paper.
Let OPT(m, n) be the expected first-best GFT.

Show that adding 2 additional sellers is sufficient to guarantee that Seller Trade Reduction achieves at least the first-best GFT of the original market. That is, show that STR(m, n+2) >= OPT(m, n) for all m, n >= 1.
\end{codebox}

\subsection{Anytime Algorithm for Prediction with Expert Advice}
\begin{codebox}
Start by reading: https://arxiv.org/pdf/2002.08994

In that paper, they establish the optimal anytime regret for two experts.

I want to understand what is the anytime regret for n experts as $n \to \infty$. Can we show that the price of anytime goes to $1$ as $n \to \infty$? Please take a look at the conjecture described in https://arxiv.org/pdf/2002.08994.

Please also have the lit reviewers do a deep search so you understand the landscape. Feel free to launch as many lit reviewers as you feel you need.
\end{codebox}

\subsection{Selling Separately vs. Bundling}
\begin{codebox}
    You are an expert theoretical computer scientist specializing in algorithmic game theory and Bayesian mechanism design. Your task is to write a formally rigorous mathematical proof.

**Problem Statement:**
Prove that for a single additive buyer with multiple items, the revenue from the better of selling the items separately or selling them as a grand bundle is a factor-3 approximation to the optimal expected revenue. The items have independent values drawn from known distributions. 

Formally, show that: $\text{OPT} \le 3 \max\{\text{SREV}, \text{BREV}\}$

**Definitions:**
- $\text{OPT}$: The expected revenue of the optimal (fully general, randomized) Bayesian Incentive Compatible mechanism.
- $\text{SREV}$: The maximum expected revenue achievable by selling each item separately at item-specific prices.
- $\text{BREV}$: The maximum expected revenue achievable by selling all items together as a single grand bundle.
- Additive Buyer: The buyer's value for a set of items is the sum of their values for the individual items.

**Reference Material:**
The following two papers are useful references:
1. *A Duality Based Unified Approach to Bayesian Mechanism Design* (Cai, Devanur, Weinberg): https://www.cs.yale.edu/homes/cai/publication/duality-journal/duality-journal.pdf
2. *A Simple and Approximately Optimal Mechanism for an Additive Buyer* (Babaioff, Immorlica, Lucier, Weinberg): https://arxiv.org/pdf/1405.6146

**Proof Outline and Instructions:**
Please think step-by-step and structure your proof carefully using LaTeX for all mathematics. If you cannot prove the factor 3 approximation, you should first try to prove that this is a factor 4 approximation, then try to strengthen  it.
\end{codebox}

\subsection{Autobidding}
We provided two prompts to the system corresponding to the $n = 2$ case and the $n \geq 2$ case.
Note that the prompts asked the system to make some mild assumption.
We removed the assumption by asking Gemini 3.1 Pro to look at the output proof and modify it accordingly.

\begin{codebox}
First, read this paper: https://arxiv.org/abs/2207.03630. Use this paper to understand the background of autobidding.

I want to consider the following mechanism. For two bidders, if their bids are $b_1$ and $b_2$,
then bidder $i$ wins with probability $b_i / (b_1 + b_2)$ and, conditional on winning, pays $b_i$.

Prove or disprove that this mechanism has a PoA of 1.5.
You can and should assume that all bidders have strictly positive value on all queries.
\end{codebox}

\begin{codebox}
First, read this paper: https://arxiv.org/abs/2207.03630. Use this paper to understand the background of autobidding.

That paper gives a mechanism that works well for two bidders. However, the setting for more than two bidders remains open.

Your goal is to first understand the lower bound in https://arxiv.org/abs/2207.03630. What kind of PoA lower bound does it give as a function of $n$ where $n$ is the number of bidders.

Once you understand that, let's try to find an upper bound. In either words, come up with a mechanism that, for all $n$, gives a PoA as close to the lower bound as you can.
For example, if the lower bound PoA is of the form $2 - \Omega(1/n)$ then an idea upper bound should be of the form $2 - O(1/n)$.
Note that your mechanism is allowed to depend on $n$ but should not depend on values, costs, etc. other than the bids.

You can and should assume that all bidders have strictly positive value on all queries.
You can and should do some literature review on what is known about this problem.
\end{codebox}

\end{document}